\documentclass[11pt]{article}

\usepackage[]{acl}

\usepackage{times}
\usepackage{latexsym}

\usepackage[T1]{fontenc}

\usepackage[utf8]{inputenc}

\usepackage{microtype}

\usepackage{inconsolata}

\usepackage{graphicx}
\usepackage{booktabs}
\usepackage{multirow}
\usepackage{amssymb}
\usepackage{amsmath}
\usepackage{algorithm}
\usepackage{algpseudocode}

\title{\textsc{RegMix-D}: Dynamic Data Mixing via Proxy Training Trajectories}

\author{Kaiyan Zhao\textsuperscript{1,2}, Zhongtao Miao\textsuperscript{1}, Akiko Aizawa\textsuperscript{2}, Yoshimasa Tsuruoka\textsuperscript{1}\\
        \textsuperscript{1}The University of Tokyo, \textsuperscript{2}National Institute of Informatics \\
        \texttt{\{kaiyan1006, miao, tsuruoka\}@logos.t.u-tokyo.ac.jp, aizawa@nii.ac.jp}}

\begin{document}
\maketitle
\begin{abstract}
Data mixture selection is critical for Large Language Model pretraining. Existing methods such as RegMix select a single static mixture by fitting a regression model on small-scale proxy runs. We propose \textsc{RegMix-D}, a simple extension of RegMix to dynamic mixing. Our key observation is that proxy runs produce not only endpoint losses, but also full loss trajectories, which can be used to further improve data mixture. By training regression model on these trajectories, we can predict optimal mixtures at multiple training stages. \textsc{RegMix-D} supports two deployment modes: an \emph{offline} variant that generates a complete mixture schedule before target training, and an \emph{online} variant that adapts the mixture during training using observed loss. Experiments on 25B tokens of the Pile dataset with a 1B parameter target model show that \textsc{RegMix-D} consistently improves over RegMix and DoReMi across 13 downstream tasks while remaining proxy-efficient: it surpasses RegMix even with only 128 proxy models (25\% of RegMix's proxy compute budget).
\end{abstract}

\section{Introduction}
Pretraining Large Language Models~(LLMs) requires assembling massive corpora from diverse domains~\citep{belenki-etal-2025-optimizing, Chen_2026}. The mixture proportions among these domains substantially affect downstream performance, making data mixture selection a critical design choice~\citep{feng2024maximizedataspotentialenhancing}. Recent work has proposed automated methods to optimize these proportions, including DoReMi~\citep{xie2023doremi}, which uses group DRO~\citep{dro} over a reference model, and RegMix \citep{liu2024regmix}, which trains a regression model on a set of small proxy models to predict loss as a function of mixture.

While effective, these methods share a common assumption: a single optimal mixture exists and can be used throughout pretraining. However, the optimal composition of different domains may shift during training~\citep{mo2025midtraininglargelanguagemodels}. Under this view, static mixtures necessarily compromise across training, leading to suboptimal performance.

Several recent methods adopt dynamic mixing to address this limitation~\citep{chen2024aioli, tikmix2025, acodm2025}. 
However, these methods introduce additional overhead on top of standard training. See Section~\ref{related_work_dynamic} for discussion.

\begin{figure*}[t]
  \centering
  \includegraphics[width=0.93\textwidth]{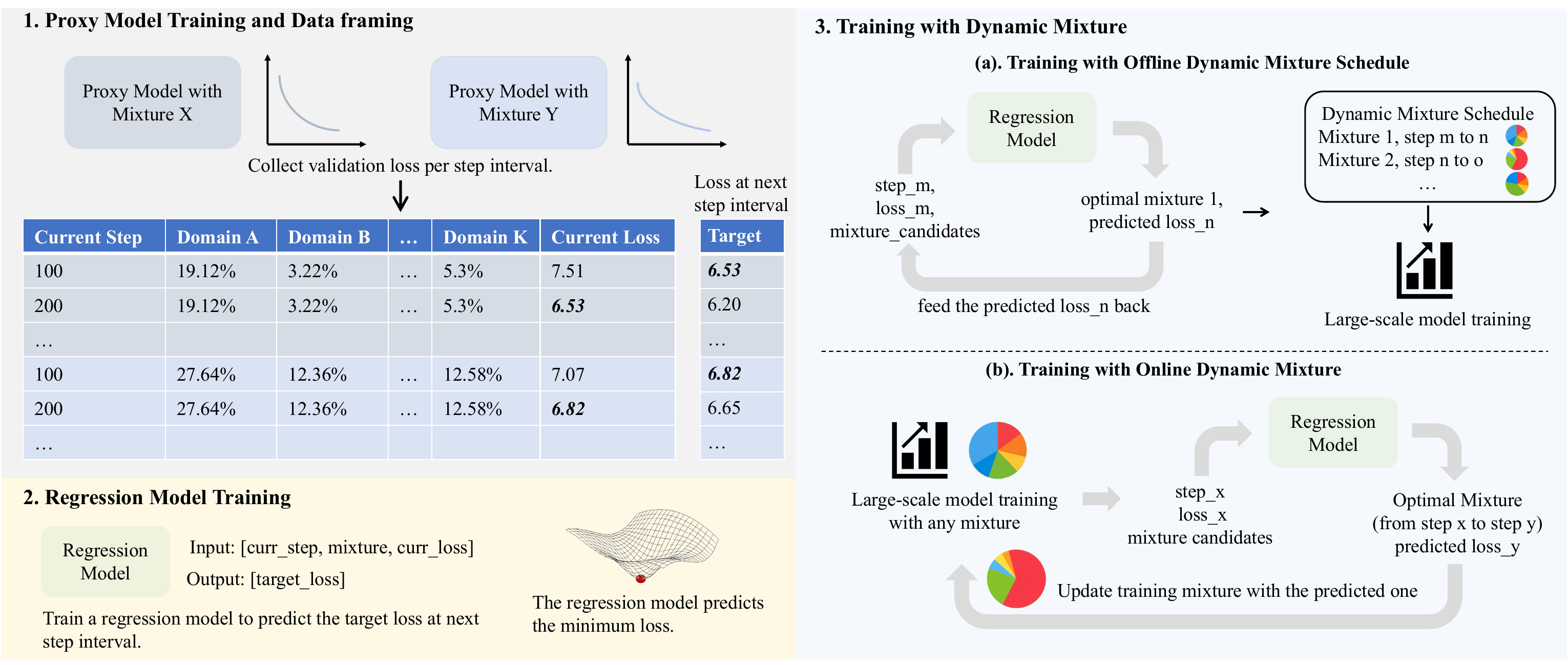}
  \caption{Overview of \textsc{RegMix-D}. We train a regression model $f$ on proxy loss trajectories (left), then deploy $f$ in two modes (right): \emph{Offline} recursively generates a complete mixture schedule before target training; \emph{Online} queries $f$ during target training using observed losses to adapt the mixture in place.}
  \label{method}

\end{figure*}

To this end, we propose \textsc{RegMix-D}, a simple extension of RegMix to dynamic mixing. Our key observation is that the loss trajectories collected during proxy runs already contain the information needed for dynamic prediction: each proxy run produces a full loss curve, not just an endpoint. We train regression models that take the current training state: $(t, m, \ell)$, where $t$ is the proxy step, $m$ is the current mixture, and $\ell$ is the current loss, to predict the loss at the next sampled step. This model can be deployed in two modes (Figure~\ref{method}): an \emph{offline} mode that recursively generates  mixture schedules before target training begins, and an \emph{online} mode that queries the regression model during training and adapts the mixture in place. Neither mode introduces additional optimization mechanisms.

We evaluate \textsc{RegMix-D} on 25B tokens of the Pile dataset~\citep{gao2020pile} with a 1B parameter target model, following the standard RegMix setup. Our contributions are as follows:
\begin{itemize}
    \item We introduce \textsc{RegMix-D}, a dynamic extension of RegMix with offline and online variants, requiring no additional training machinery beyond the existing regression framework.
    \item Across all tested granularities, \textsc{RegMix-D} consistently outperforms RegMix on validation loss and achieves superior performance across 13 downstream tasks.
    \item \textsc{RegMix-D} is data-efficient: 128 proxy models (25\% of RegMix's compute budget) suffice to surpass RegMix with 512 proxy models.
\end{itemize}

\section{Related Work}
\label{related_work_dynamic}
\paragraph{Static Data Mixing.}
Conventional methods select a single mixture for the whole training. DoReMi \citep{xie2023doremi} uses group DRO with a reference model to upweight domains with high excess loss. DoGE \citep{pmlr-v235-fan24-doge} reweights domains based on gradient-level generalization contribution. RegMix \citep{liu2024regmix} trains small proxy models with diverse mixtures and fits a regression model that predicts validation loss from mixture candidates, then selects the optimal mixture by search. An alternative line of work selects mixtures via weighted model merging rather than direct proxy training~\citep{li2026decouple, wang2026mergemixoptimizingmidtrainingdata}. \textsc{RegMix-D} builds directly on RegMix's regression framework, generalizing single-mixture selection to time-varying schedules.

\paragraph{Dynamic Data Mixing}

Several recent methods adopt time-varying mixtures. Aioli \citep{chen2024aioli} estimates mixing-law parameters online during target training and updates proportions via exponentiated gradient descent. TiKMiX \citep{tikmix2025} reweights domains using group gradient influence on a target distribution. AC-ODM \citep{acodm2025} trains an actor-critic network on proxy models and transfers it to target training.
\textsc{RegMix-D} differs from them along three axes: (i) \emph{No additional machinery during target training}: existing dynamic methods introduce online optimization, gradient-influence computation, or RL queries on top of target training, whereas \textsc{RegMix-D} confines all optimization to the proxy phase. (ii) \emph{Reuses RegMix's signal}: we predict from proxy loss trajectories already produced in the RegMix pipeline, requiring no new instrumentation. (iii) \emph{Two deployment modes}: a single regression model supports both offline and online adaptation. 
\section{\textsc{RegMix-D}}
\label{sec:method}
\subsection{Preliminaries}
RegMix~\citep{liu2024regmix} selects a single static data mixture by training $M$ small \emph{proxy} models (one per sampled mixture) for a fixed number of steps $T_p$, recording the validation loss at the final step, and fitting a regression model that maps a mixture to its predicted validation loss. The optimal mixture is then identified by searching over candidate mixtures and selecting those with the lowest predicted loss. The \emph{target} model is subsequently trained from scratch with this single mixture for the entire run. This discards two sources of information in proxy runs: (i) intermediate losses along each proxy trajectory, and (ii) how the optimal mixture might evolve across training.


\subsection{Trajectory-Conditioned Regression}
We log validation loss at multiple step intervals during training instead of only the endpoint loss. Given a desired number of switch points $N$ (at which the mixture would change), we select $N+1$ intervals $\{t_1, t_2, \ldots, t_{N+1}\}$ from the proxy run. For each proxy run with mixture $m^{(i)}$, this yields a trajectory $\{\ell_1^{(i)}, \ldots, \ell_{N+1}^{(i)}\}$ of validation losses.

We construct training pairs for a new regression model $f_N$ over consecutive intervals as shown in Figure~\ref{method}, Part 1. For each proxy $i$ and each adjacent pair $(t_\ell, t_{\ell+1})$, we form one training example:
\begin{equation}
\underbrace{\big(t_\ell,\; m^{(i)},\; \ell_\ell^{(i)}\big)}_{\text{input}} \;\longrightarrow\; \underbrace{\ell_{\ell+1}^{(i)}}_{\text{target}}.
\label{eq:training_pair}
\end{equation}
That is, $f_N$ learns to predict the validation loss at the next step interval, conditioned on (a) the current proxy step $t_\ell$, (b) the current mixture $m^{(i)}$, and (c) the current observed loss $\ell_\ell^{(i)}$. This produces $M \times N$ training pairs in total. Crucially, $f_N$ predicts a \emph{local} transition $(t_\ell, m, \ell_\ell) \rightarrow \ell_{\ell+1}$ rather than a global mixture-to-endpoint map. This locality is what enables dynamic scheduling: at any training state, we can ask $f_N$ which mixture is expected to drive the loss lowest in the next interval. Note that $f_N$ is specific to a given $N$, since the training pairs depend on the interval numbers.

\subsection{Offline Mixture Schedule (Algorithm~\ref{alg:offline})}
Given the regression model $f_N$ trained at the desired $N$, we can generate mixture schedules with $N$ switch points $\{s_1, s_2, \ldots, s_N\}$, corresponding to the step intervals used to train $f_N$. The schedule divides target training into $N+1$ segments. The first segment uses the human-curated Pile prior $m_0^{\text{human}}$, reflecting the assumption that no prior knowledge about the target corpus is available at the start of training, and each subsequent segment uses a regression-predicted mixture.

Starting from an initial loss $\ell_0$, we iterate over switch points $s_1, \ldots, s_N$. At each switch point $s_j$, we search for the mixture $m_j^*$ minimizing $f_N(s_j, m, \ell_{j-1})$ over Dirichlet-sampled candidates, record $(s_j, m_j^*)$ as one schedule entry, and update $\ell_j \leftarrow f_N(s_j, m_j^*, \ell_{j-1})$ as the input loss for the next iteration. The novelty lies in how $f_N$ is queried across the $N$ switch points: predictions are made \emph{recursively}, with each prediction's output loss fed into the next prediction as input. The final schedule $\{(s_j, m_j^*)\}_{j=1}^{N}$ is mapped to target training via $s_j^{\text{target}} = s_j \cdot (T_{\text{target}} / T_p)$, where $T_{\text{target}}$ stands for the training steps of the target model.


\subsection{Online Dynamic Mixture (Algorithm~\ref{alg:online})}
\label{sec:online}


The recursive use of predictions in offline mode may iteratively accumulate error. The online variant replaces predicted losses with \emph{observed} target-model losses at each switch point, grounding subsequent mixture decisions during target training.

Concretely, the same regression model $f_N$ trained on proxy trajectories is loaded as a frozen predictor at the start of target training. At each switch point $S_j = s_j \cdot (T_{\text{target}}/T_p)$, we query $f_N$ using the validation loss $\hat{\ell}_{S_j}$ measured on the target model, then select the next segment's mixture via the same Dirichlet-sampling and top-$k$ averaging used in the offline variant.

\paragraph{Cross-scale loss correction.}
Because $f_N$ is trained on proxy-scale losses (proxy model size $P_p$) while target losses come from a much larger model (size $P_t$), the two are not directly comparable. We apply a power-law correction to map target losses to proxy scale before invoking $f_N$~\citep{kaplan2020scalinglawsneurallanguage}:
\begin{equation}
\tilde{\ell}_{S_j} \;=\; \hat{\ell}_{S_j} \cdot \bigl(P_t / P_p\bigr)^{\beta},
\label{eq:loss_scale}
\end{equation}
where $\beta$ is a small positive constant. In our setup $P_t/P_p = 1000$, and we use $\beta = 0.05$, selected from a sweep over $\beta \in \{0.01, 0.02, 0.05, 0.10\}$; final downstream performance varies by less than $0.25$ across this range (Table~\ref{tab:ablation}), indicating that online adaptation is robust to the precise choice of $\beta$. The query at switch point $S_j$ thus becomes $f_N(s_j,\, m,\, \tilde{\ell}_{S_j})$.

\section{Experiments}
\begin{table*}[t]
\centering
\small
\setlength{\tabcolsep}{4pt}
\resizebox{0.65\textwidth}{!}{
\begin{tabular}{lccccccc}
\toprule
\multirow{2.5}{*}{\textbf{Task}} & \multirow{2.5}{*}{\textbf{Human}} & \multirow{2.5}{*}{\textbf{DoReMi}} & \multirow{2.5}{*}{\textbf{RegMix}} & \multicolumn{2}{c}{\textbf{RegMix-D (128)}} & \multicolumn{2}{c}{\textbf{RegMix-D (512)}} \\
\cmidrule(lr){5-6} \cmidrule(lr){7-8}
 & & & & Offline & Online & Offline & Online \\
\midrule
HellaSwag      & 37.93	& 43.01	& 43.44 & \underline{45.65} & 45.45  & 45.24 & \textbf{45.68} \\
PIQA           & 64.99	& 68.05	& 68.61 & 69.16 & \underline{69.23}  & 69.11 & \textbf{69.75} \\
OpenBookQA     & 29.17	& 30.70	& 30.33 & 30.03 & 31.20  & \textbf{32.00} & \underline{31.77} \\
LAMBADA        & 27.09	& 31.58	& 32.68 & 32.63 & \textbf{34.45}  & 34.23 & \underline{34.43} \\
SciQ           & 79.67	& 81.42	& 80.98 & 83.35 & \textbf{83.67}  & 82.53 & \underline{83.45} \\
ARC-Easy       & 49.06	& 50.36	& 51.16 & 53.12 & \underline{53.93}  & 53.49 & \textbf{54.27} \\
ARC-Challenge  & 25.61	& 25.78	& 26.39 & \underline{27.23} & 26.55  & 26.32 & \textbf{27.57} \\
COPA           & 65.00	& 67.50	& 68.33 & 70.17 & 70.17  & \underline{71.00} & \textbf{71.17} \\
RACE           & 29.87	& 31.07	& 30.75 & 31.21 & 30.91  & \underline{31.83} & \textbf{32.06} \\
LogiQA         & 25.47	& \textbf{26.93} & \underline{26.37} & 26.14 & 25.60  & 25.35 & 24.86 \\
QQP            & 42.48	& 50.50	& \textbf{51.58} & 50.67 & \underline{51.53}  & 49.95 & 50.80 \\
WinoGrande     & 51.87	& 52.09	& \underline{52.80} & \textbf{53.46} & 51.72  & 52.03 & 52.14 \\
MultiRC        & \textbf{54.15}	& \underline{53.31}	& 52.86 & 50.25 & 51.57  & 52.13 & 51.71 \\
\midrule
Avg   & 44.80	& 47.10	& 47.41 & 47.93 & \underline{48.15} & 48.09 & \textbf{48.44} \\
\bottomrule
\end{tabular}}
\caption{Downstream performance on 13 tasks, averaged over 0-5 shot settings. All methods train a 1B target model on 25B tokens. RegMix-D variants use $N=5$ switch points. \textbf{Bold} and \underline{underline} indicate the best/second best result per row. RegMix-D (128) uses 128 proxy models, approximately 25\% of RegMix's 512-proxy compute budget.}
\label{tab:main_results}
\end{table*}

\subsection{Experimental Setup}
Following~\citet{liu2024regmix}, we use the 1B TinyLlama~\citep{zhang2024tinyllama} trained on 25B Pile~\citep{gao2020pile} dataset (17 domains) as the target model and a 1M parameter TinyLlama variant as the proxy model in the main experiments. 
We compare against three baselines: Human, DoReMi~\citep{xie2023doremi}, and RegMix~\citep{liu2024regmix}. 
We report \textbf{RegMix-D (128)} and \textbf{RegMix-D (512)}, and two deployment variants for each: \textbf{offline} and \textbf{online} with $N=5$ switch points in main experiments. 
Average results for 13 tasks from \texttt{lm-eval-harness}\footnote{\url{https://github.com/EleutherAI/lm-evaluation-harness}} \citep{eval-harness} are reported. More details can be found in Appendix~\ref{app:imp}.

\subsection{Main Results}
Table~\ref{tab:main_results} reports downstream performance, where \textsc{RegMix-D} consistently outperforms baselines at both proxy budgets. Notably, even with only 128 proxies, 25\% of the proxy compute used by RegMix, both \textsc{RegMix-D} variants still surpass RegMix in average score. Figure~\ref{loss_curve} shows the pile-cc validation loss across training: all four \textsc{RegMix-D} variants achieve consistently lower loss than RegMix. The corresponding dynamic mixtures differ substantially from RegMix's static mixture, and we visualize the pile-cc weight trajectory in Figure~\ref{pilecc}, Appendix~\ref{app:mixture_viz}.

\textbf{Online generally outperforms offline.} Across both proxy budgets, the online variant achieves higher average accuracy than the offline variant. We attribute this to the gap between proxy and target dynamics: offline schedules are generated entirely from proxy loss trajectories, while online predictions condition on the actual target model's observed loss at each switch point. This grounding in real target dynamics provides a more accurate signal for downstream performances.

\textbf{Improvements are broad rather than concentrated.} \textsc{RegMix-D} variants achieve top-1 or top-2 performance on 11 of 13 tasks. We do not observe a single task category (e.g., commonsense, reading comprehension) where \textsc{RegMix-D} systematically wins or loses, suggesting the gains stem from a generally better mixture rather than implicit specialization toward particular tasks.

\textbf{Compute scaling helps modestly.} Increasing proxies from 128 to 512 yields a smaller average gain. This suggests that the dynamic-vs-static gap dominates the proxy-count gap in our setup: RegMix-D extracts more signal from each proxy run than RegMix does, by exploiting the full loss trajectory rather than a single endpoint. We further analyze proxy count in the next section.

\subsection{Ablation Study}

Table~\ref{tab:ablation}, Appendix, reports ablations on the three main design choices of \textsc{RegMix-D}. \textbf{All 26 configurations across the three studies outperform the RegMix baseline (47.41)}, indicating that the benefit of dynamic mixing is robust to specific hyperparameter choices rather than reliant on careful tuning. Three more specific findings follow. First, the number of switch points $N$ has a moderate but consistent effect: $N{=}5$ is best in all four proxy/variant settings, while $N{=}3,7,9$ also reliably surpass RegMix. Second, the cross-scale loss correction factor $\beta$ has limited influence on the Online variant, with maximum Avg variation of $0.25$. This robustness justifies treating $\beta$ as a fixed constant rather than a tuned hyperparameter. Third, increasing the proxy model size from 1M to 120M yields virtually identical performance (within $0.05$ Avg for both variants), a consistent finding with~\citet{liu2024regmix}. We discuss the overhead of \textsc{RegMix-D} in Appendix~\ref{overhead}.

\section{Conclusion}
We present \textsc{RegMix-D}, a simple regression-based extension of \textsc{RegMix} that generates dynamic data mixture ratios, with both offline and online variants and no additional optimization machinery. On the Pile dataset with a 1B-parameter target model, \textsc{RegMix-D} improves over \textsc{RegMix} and DoReMi across 13 downstream tasks, and surpasses \textsc{RegMix} even with 25\% of its proxy compute. Ablation studies show the gains are robust to switch-point count, scaling factor, and proxy size, indicating that effective dynamic mixing is achievable at low cost.

\clearpage
\section*{Limitations}

\textsc{RegMix-D} has several limitations that motivate future work. First, our experiments use a single target-domain validation loss (pile-cc) as the optimization signal, following RegMix's setup. Extending \textsc{RegMix-D} to multi-target or domain-agnostic objectives requires a non-trivial extension of the regression framework and is left to future work.
Second, while several recent dynamic data mixing methods (e.g., Aioli, TiKMiX) report strong results in their respective settings, fair head-to-head comparison is challenging: existing works use different corpora (e.g., SlimPajama vs.\ Pile), domain partitions, target model scales, and downstream benchmark suites, making direct numerical comparison costly to reproduce within a single setup. We therefore restrict our quantitative comparisons to RegMix and DoReMi, which share \textsc{RegMix-D}'s deployment setup.

\section*{Ethnical Statements}
This work studies data mixture selection for language model pretraining. We use the Pile~\citep{gao2020pile}, a publicly released corpus, and follow its existing domain partitioning. \textsc{RegMix-D} does not introduce new training data, generation capabilities, or downstream applications beyond those already present in standard pretraining pipelines, and we therefore do not foresee direct ethical risks specific to our method. We use LLMs only for language and grammar polishing.

\bibliography{custom}

\appendix

\section{Appendix}
\label{sec:appendix}

\subsection{Implementation Details}
\label{app:imp}
The hyperparameters we used are: AdamW optimizer~\citep{loshchilov2018decoupled} with weight decay 0.1, learning rate 4e-4, context length 2048, and global batch size 512 (achieved via gradient accumulation). Proxy models are trained on 1 H800 GPU for 1{,}000 steps (1M tokens per step) and the target model is trained on 8 GPUs for 25{,}000 steps, totaling 25B tokens. Human stands for the original Pile token distribution. The adjusted  17 domains are from~\citet{liu2024regmix}.

For all baselines (Human, DoReMi, and RegMix), we use the open-sourced mixtures released and train target models under our exact hyperparameter setup. This isolates the contribution of the mixture itself from any unrelated training-recipe differences. All target model runs use a single seed. We use LightGBM \citep{lightgbm} as the regression model, matching the choice in RegMix. 

We use the Pile-CC validation loss as the target predicted loss. Predictions over candidate mixtures are made by Dirichlet sampling 100K candidates and averaging the top-128 lowest predicted validation loss.

The full list of our evaluated tasks: HellaSwag~\citep{zellers-etal-2019-hellaswag}, PIQA~\citep{Bisk2020piqa}, OpenBookQA~\citep{OpenBookQA2018}, LAMBADA~\citep{paperno-etal-2016-lambada}, SciQ~\citep{Welbl2017sciq}, ARC-Easy~\citep{Clark2018arc}, ARC-Challenge~\citep{Clark2018arc}, COPA~\citep{sarlin20supergluecopa}, RACE~\citep{lai-etal-2017-race}, LogiQA~\citep{liu2020logiqa}, QQP~\citep{wang-etal-2018-glue-qqp}, WinoGrande~\citep{sakaguchi2019winogrande}, and MultiRC~\citep{khashabi-etal-2018-looking-multirc}.

\begin{algorithm}[t]
\caption{Offline Mixture Schedule Generation}
\label{alg:offline}
\begin{algorithmic}[1]
\Require regression model $f_N$, switch points $\{s_1, \ldots, s_N\}$, initial loss $\ell_0 = \frac{1}{M}\sum_{i=1}^{M} \ell_1^{(i)}$, candidate count $C$, top-$k$, human prior $m_0^{\text{human}}$, concentration $\alpha$
\Ensure schedule $\mathcal{S} = \{(s_j, m_j^*)\}_{j=1}^{N}$
\State $\mathcal{S} \gets \emptyset$
\For{$j = 1, \ldots, N$}
    \State Sample $\{m^{(c)}\}_{c=1}^{C} \sim \mathrm{Dir}(\alpha \cdot m_0^{\text{human}})$
    \State $\hat{\ell}^{(c)} \gets f_N(s_j, m^{(c)}, \ell_{j-1})$ \quad for $c=1,\ldots,C$
    \State $\mathcal{I} \gets \mathrm{TopK}_{c}\bigl(-\hat{\ell}^{(c)}\bigr)$ \Comment{indices of $k$ lowest losses}
    \State $m_j^* \gets \frac{1}{k}\sum_{c \in \mathcal{I}} m^{(c)}$
    \State $\ell_j \gets f_N(s_j, m_j^*, \ell_{j-1})$ \Comment{input loss for next iter}
    \State $\mathcal{S} \gets \mathcal{S} \cup \{(s_j, m_j^*)\}$
\EndFor
\State \Return $\mathcal{S}$
\end{algorithmic}
\end{algorithm}

\begin{algorithm}[h]
\caption{Online Dynamic Mixture}
\label{alg:online}
\begin{algorithmic}[1]
\Require regression model $f_N$, target model $\mathcal{M}$, target-step switch points $\{S_1, \ldots, S_N\}$, candidate count $C$, top-$k$, scaling exponent $\beta$, size ratio $P_t/P_p$, human prior $m_0^{\text{human}}$, concentration $\alpha$
\State Initialize $\mathcal{M}$ with mixture $m_0^{\text{human}}$
\For{each training step $t$}
    \State Take one training step of $\mathcal{M}$
    \If{$t \bmod T_{\text{eval}} = 0$}
        \State $\hat{\ell}_t \gets \mathrm{validate}(\mathcal{M})$ \Comment{cached for online use}
    \EndIf
    \If{$t \in \{S_1, \ldots, S_N\}$, say $t = S_j$}
        \State $\tilde{\ell}_{S_j} \gets \hat{\ell}_{S_j} \cdot (P_t/P_p)^{\beta}$ \Comment{scale to proxy}
        \State $s_j \gets t \cdot T_p / T_{\text{target}}$ \Comment{proxy-step image}
        \State Sample $\{m^{(c)}\}_{c=1}^{C} \sim \mathrm{Dir}(\alpha \cdot m_0^{\text{human}})$
        \State $\hat{\ell}^{(c)} \gets f_N(s_j, m^{(c)}, \tilde{\ell}_{S_j})$ \quad for $c=1,\ldots,C$
        \State $\mathcal{I} \gets \mathrm{TopK}_{c}\bigl(-\hat{\ell}^{(c)}\bigr)$
        \State $m_j^* \gets \frac{1}{k}\sum_{c \in \mathcal{I}} m^{(c)}$
        \State Update $\mathcal{M}$'s training mixture to $m_j^*$
    \EndIf
\EndFor
\end{algorithmic}
\end{algorithm}

\begin{figure}[t]
  \centering
  \includegraphics[width=0.49\textwidth]{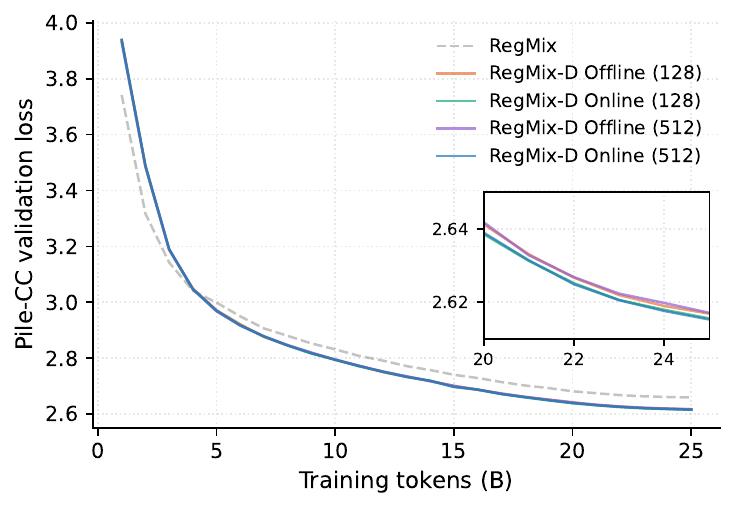}
  \caption{Pile-CC validation loss across training for RegMix and all four \textsc{RegMix-D} variants (Offline/Online $\times$ 128/512 proxies), at $N{=}5$ switch points. All four \textsc{RegMix-D} variants achieve consistently lower validation loss than RegMix throughout training. The inset zooms into the final 5B tokens, where RegMix-D Online (512) attains the lowest loss.}
\label{fig:loss_curves}
  \label{loss_curve}

\end{figure}

\subsection{Overhead for \textsc{RegMix-D}}
\label{overhead}
\textsc{RegMix-D}'s online adaptation adds minimal cost to target training. Table~\ref{tab:wallclock} reports wall-clock training time on 8 GPUs (averaged over 3 runs): \textsc{RegMix-D} Offline matches RegMix exactly (the schedule is generated before training), while \textsc{RegMix-D} online incurs only $+0.37\%$ overhead from per-switch-point regression queries. This confirms that the additional flexibility of online adaptation comes at negligible computational cost.
\begin{table}[h]
\centering
\small
\setlength{\tabcolsep}{8pt}
\begin{tabular}{lcc}
\toprule
\textbf{Method} & \textbf{Time (hours)} & \textbf{Overhead} \\
\midrule
RegMix             & 21.77 & --- \\
RegMix-D (Offline) & 21.76 & $\approx 0\%$ \\
RegMix-D (Online)  & 21.85 & $+0.37\%$ \\
\bottomrule
\end{tabular}
\caption{Target-model training wall-clock time on 8 GPUs (25{,}000 steps, averaged over 3 runs).}
\label{tab:wallclock}
\end{table}

As for the proxy model training, one 1M proxy model requires about 0.5 hours on a single H800, reducing proxy number from 512 to 128 thus would reduce the total proxy training cost from approximately 256 to 64 H800 GPU-hours.

\subsection{From Proxy Timescale to Target Training}
\textsc{RegMix-D}'s regression model is trained entirely on proxy-scale trajectories ($T_p$ steps), while target training is much longer ($T_{\text{target}} \gg T_p$. We assume that proxy and target training share enough loss-curve structure to make the proxy-predicted switch points transferable. Concretely, each proxy switch point $s_j$ is linearly mapped to a target step $s_j \cdot (T_{\text{target}}/T_p)$. 
This assumption underlies RegMix's original design as well: the static mixture selected from proxy losses is similarly assumed to transfer to target training. \textsc{RegMix-D} inherits this assumption without strengthening it.

\begin{table*}[t]
\centering
\small
\setlength{\tabcolsep}{6pt}

\begin{minipage}[t]{0.40\textwidth}
\centering
\textbf{(a) Number of switch points $N$} \\[2pt]
\begin{tabular}{lcccc}
\toprule
& \multicolumn{2}{c}{\textbf{128 proxies}} & \multicolumn{2}{c}{\textbf{512 proxies}} \\
\cmidrule(lr){2-3} \cmidrule(lr){4-5}
$N$ & Off. & On. & Off. & On. \\
\midrule
3 & 47.47 & 48.13 & 47.55 & 48.01 \\
5 & \textbf{47.93} & \textbf{48.15} & \textbf{48.09} & \textbf{48.44} \\
7 & 47.78 & 47.74 & 47.84 & 47.98 \\
9 & 47.58 & 47.82 & 48.01 & 48.40 \\
\bottomrule
\end{tabular}
\end{minipage}
\hfill
\begin{minipage}[t]{0.25\textwidth}
\centering
\textbf{(b) Scaling factor $\beta$} \\[2pt]
\begin{tabular}{lcc}
\toprule
$\beta$ & \textbf{128} & \textbf{512} \\
\midrule
0.01 & 48.08 & 48.24 \\
0.02 & \textbf{48.24} & 48.42 \\
0.05 & 48.15 & \textbf{48.44} \\
0.10 & 47.99 & 48.27 \\
\bottomrule
\end{tabular}
\end{minipage}
\hfill
\begin{minipage}[t]{0.30\textwidth}
\centering
\textbf{(c) Proxy size} \\[2pt]
\begin{tabular}{lcc}
\toprule
Size & \textbf{Offline} & \textbf{Online} \\
\midrule
1M   & 47.93 & 48.15 \\
60M  & \textbf{47.97} & 48.14 \\
120M & 47.93 & \textbf{48.19}\\
\bottomrule
\end{tabular}
\end{minipage}

\caption{Ablations on \textsc{RegMix-D} (downstream Avg, \%). All configurations exceed the RegMix baseline (47.41). \textbf{(a)} Number of switch points $N$ at the 128- / 512-proxy setting. \textbf{(b)} Cross-scale loss correction factor $\beta$ for the Online variant; maximum variation is $0.25$. We use $\beta{=}0.05$ in main experiments as default. \textbf{(c)} Proxy model size analysis at the 128-proxy setting. \textbf{Bold}: best per row.}
\label{tab:ablation}
\end{table*}

\subsection{Visualization of the Dynamic Mixture}
\label{app:mixture_viz}
\begin{figure}[t]
  \centering
  \includegraphics[width=0.49\textwidth]{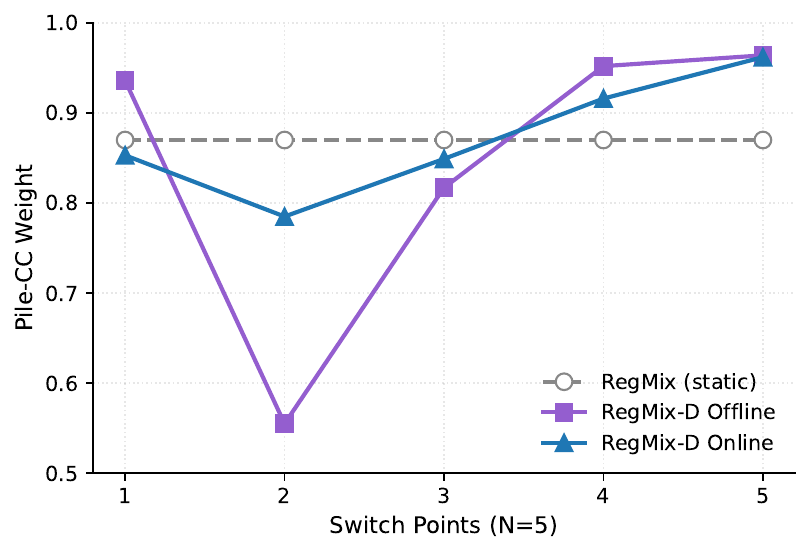}
  \caption{Pile-CC weight across training process for RegMix (static, dashed) and \textsc{RegMix-D} (dynamic, $N{=}5$, 512 proxies). The \textsc{RegMix-D} variants assign substantially different mixture weights than RegMix. This illustrates the dynamic schedules discovered by our method.}
  \label{pilecc}

\end{figure}
Figure~\ref{pilecc} visualizes the pile-cc weights produced by \textsc{RegMix-D} (Offline and Online) at the $N{=}5$, 512-proxy setting, compared with RegMix's static weight of 0.870. The dynamic schedules differ markedly from the static baseline: both variants assign substantially different pile-cc weights in 4 of 5 segments. The Offline variant exhibits a more pronounced dip in segment 2 (0.555), reflecting a stronger early redistribution of weight toward non-dominant domains, before concentrating back to high pile-cc weights in later segments. The Online variant follows a similar overall trajectory but with smaller fluctuations, likely because it conditions on actually observed target-model losses rather than recursively predicted ones.

We restrict the visualization to pile-cc because the remaining 16 Pile domains each contribute less than 5\% of the mixture, and their individual weight changes are visually difficult to distinguish on a shared axis. 

\end{document}